\documentclass[10pt,twocolumn,letterpaper]{article}

\usepackage{iccv}
\usepackage{times}
\usepackage{epsfig}
\usepackage{graphicx}
\usepackage{amsmath}
\usepackage{amssymb}
\usepackage{algorithm}
\usepackage{bm}
\usepackage{algorithmic}
\usepackage{listings}

\usepackage{multirow}
\usepackage{booktabs}
\usepackage{array}
\usepackage{subcaption}
\usepackage{colortbl}
\usepackage{multirow}
\usepackage{colortbl}
\usepackage[accsupp]{axessibility}

\usepackage[dvipsnames]{xcolor}
\definecolor{OceanBlue}{RGB}{0, 119, 190}
\definecolor{ForestGreen}{RGB}{34, 139, 34}
\definecolor{IndigoPurple}{HTML}{6A5ACD}
\definecolor{DarkIndigoPurple}{HTML}{4B3B9B}
\definecolor{tabhighlight}{HTML}{f2f2f2}
\definecolor{indianred}{RGB}{205,92,92}
\definecolor{royalblue}{RGB}{65,105,225}
\usepackage{pifont}

\usepackage[pagebackref=true,breaklinks=true,colorlinks,bookmarks=false,
citecolor=IndigoPurple,linkcolor=OceanBlue]{hyperref}
\usepackage[capitalise]{cleveref}

\crefname{table}{Tab.}{Tabs.}
\newcommand{\tableCellHeight}{1}
\newcommand{\tabstyle}[1]{
  \setlength{\tabcolsep}{#1}
  \renewcommand{\arraystretch}{\tableCellHeight}
  \centering
  \small
}
\newcommand{\xmark}{\ding{55}}

\newcolumntype{g}{>{\columncolor{tabhighlight}}c}

\iccvfinalcopy %

\def\iccvPaperID{5321} %

\begin{document}

\title{Black Box Few-Shot Adaptation for Vision-Language models}

\author{Yassine Ouali\textsuperscript{\rm 1}
\quad Adrian Bulat\textsuperscript{\rm 1}
\quad Brais Matinez\textsuperscript{\rm 1}
\quad Georgios Tzimiropoulos\textsuperscript{\rm 1,\rm 2}\\
\textsuperscript{\rm 1}Samsung AI Cambridge \quad \textsuperscript{\rm 2}Queen Mary University of London\\
{\tt\footnotesize \{y.ouali, brais.a\}@samsung.com}, {\tt\small adrian@adrianbulat.com}, {\tt\small g.tzimiropoulos@qmul.ac.uk}
}

\maketitle
\ificcvfinal\thispagestyle{empty}\fi

\begin{abstract}
Vision-Language (V-L) models trained with contrastive learning to align the visual and language modalities
have been shown to be strong few-shot learners. Soft prompt learning is the method of choice for few-shot
downstream adaptation aiming to bridge the modality gap caused by the distribution shift induced by the new domain.
While parameter-efficient, prompt learning still requires access to the model weights and can be computationally
infeasible for large models with billions of parameters. To address these shortcomings, in this work, we describe
a black-box method for V-L few-shot adaptation that (a) operates on pre-computed image and text features and hence
works without access to the model's weights, (b) it is orders of magnitude faster at training time, (c) it is amenable
to both supervised and unsupervised training, and (d) it can be even used to align image and text features computed from
uni-modal models. To achieve this, we propose Linear Feature Alignment (LFA), a simple linear approach for V-L re-alignment
in the target domain. LFA is initialized from a closed-form solution to a least-squares problem and then it is iteratively
updated by minimizing a re-ranking loss. Despite its simplicity, our approach can even surpass soft-prompt learning methods
as shown by extensive experiments on 11 image and 2 video datasets.\\
Code available at: {\footnotesize{\url{https://github.com/saic-fi/LFA}}}
\end{abstract}

\section{Introduction}
\label{sec:intro}

\begin{figure*}[t]
  \centering
  \includegraphics[width=1.0\textwidth]{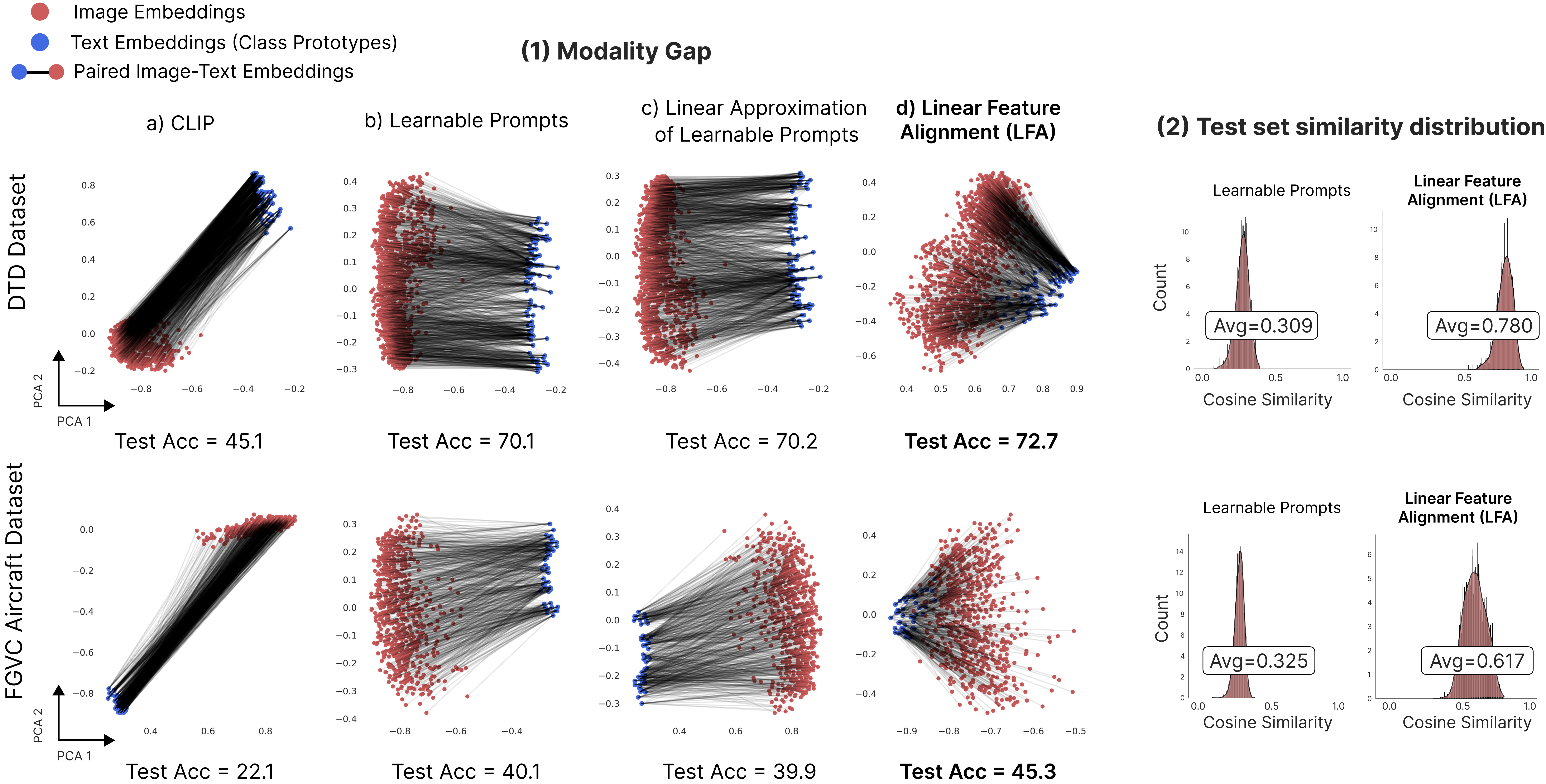}
  \caption{\textbf{Effect of Linear Feature Alignment (LFA):} We use 16-shot (per class) 
  training data for two fine-grained image classification datasets: DTD and FGVC Aircraft.
  In \textbf{(1)}, we show the training set modality gap between paired
  \textcolor{indianred}{image embeddings} and \textcolor{royalblue}{class prototypes}
  following the same procedure as in \cite{liang2022mind},
  and the obtained test set accuracy. 
  The embeddings are visualized in 2D using PCA.
  (a) With CLIP features, we observe
  a big modality gap, resulting in low test accuracy.
  (b) After learning a set of soft-prompts, we obtain a better alignment and improved results.
  However, the modality gap is still not sufficiently reduced.
  (c) A simple linear transformation $\mathbf{W}$ that maps the original class
  prototypes (obtained using only the class names) to the ones obtained with soft-prompt learning induces a similar modality gap.
  (d) Motivated by (c) we propose LFA, which aligns the image embeddings with
  their class prototypes via linear mapping $\mathbf{W}$, obtained by solving a
  simple optimization problem. LFA results in better alignment and improved accuracy.
  In \textbf{(2)}, we show that with LFA, the test image features are closely aligned with
  their corresponding class prototypes, resulting in higher cosine similarity
   scores compared to the ones obtained with soft prompts.}
  \vspace{-0.15in}
  \label{fig:intro}
\end{figure*}

Large-scale Vision-Language (V-L) models~\cite{radford2021learning} trained with contrastive learning currently represent the de-facto approach for few-shot visual adaptation. Their unprecedented success lies in part in the strength of the joint V-L embedding space learned by aligning the image and text modalities. However, when a V-L model is applied to a new domain, the domain shift exacerbates the V-L modality gap~\cite{liang2022mind}, and some sort of adaptation is required to obtain high accuracy (see \cref{fig:intro}(a)). The question that we want to address in this paper is: ``can we effectively adapt a V-L model to a new domain by having access to pre-computed features only?'' We call this black-box adaptation. 

Similar to their NLP counterparts~\cite{radford2021learning,lester2021power}, soft prompt learning
has emerged as the preferred technique for adapting a V\&L to new tasks. Specifically, a number of
works~\cite{zhou2022learning,zhou2022conditional,bulat2022language,zhu2022prompt,
chen2022prompt,shu2022test,lu2022prompt,huang2022unsupervised} have proposed to replace the manually
designed prompts of~\cite{radford2021learning} (\eg, \texttt{a photo of a \{cls\_name\}}), with a sequence
of learnable vectors, coined \textit{soft prompts}. These are passed as input to the text encoder jointly
with the class name \texttt{cls\_name} to create the new prototypes effectively reducing the modality gap.
The prompts are learned in a supervised manner using a standard cross-entropy loss given a set of labeled images. 

While soft-prompt learning approaches demonstrate promising results on various downstream
tasks~\cite{zhou2022learning,zhou2022conditional,bulat2022language}, they suffer from two
limitations: (1) They require access to the model's weights, and (2) the training cost can
be prohibitive, especially on commodity hardware and low-power devices, as computing the gradients
and updating the prompts for thousands of iterations~\cite{zhou2022learning} is required.
As the model's size continues to grow (\eg, billion-parameter models such as
CoCa \cite{yu2022coca}), and the industry transitions to models as a service (\eg, via API),
the existing  methods can be rendered either inapplicable or impractical.

In this work, we seek to address these limitations by bridging the modality gap directly in the feature space
without prompting or access to the model's weights. We first empirically show that a simple linear
transformation can approximate the alignment effect of prompt
learning (\eg, see \cref{fig:intro} and Sec.~\ref{sec:approximating_prompting}).
Importantly, this shows that it is possible to derive a black-box method that manipulates the CLIP features directly for
downstream adaptation. Then, motivated by this observation, we propose Linear Feature Alignment (LFA), a black-box method
that learns a linear mapping $\mathbf{W}$, obtained by solving a simple optimization problem, which effectively aligns the
image features $\mathbf{X}$ with their text class prototypes $\mathbf{Y}$,
\ie, $\mathbf{X} \xrightarrow{\mathbf{W}} \mathbf{Y}$. Specifically, our contributions are:

\begin{itemize}
\setlength{\itemsep}{-0.2em}
\item We propose the very first black-box method for the few-shot adaptation of V-L models.
\item To this end, and motivated by the observation that prompting can be successfully approximated by a
linear transformation, we propose Linear Feature Alignment (LFA), an efficient and effective adaptation method for reducing the modality gap
between the image and text modalities of a V-L model. LFA is initialized by $\beta$-Procrustes,
a regularized version of orthogonal Procrustes, and then minimizes a simple adaptive reranking loss adapted for V-L models.
\item We propose both supervised and unsupervised formulations for LFA and moreover, a variant that
works for the case of base-to-new (\ie, zero-shot) generalization.
\item We demonstrate that LFA can achieve better alignment (\eg, see \cref{fig:intro} (1d) and (2)) and improved
accuracy compared to prompt learning methods while being more efficient (\ie, training takes few minutes)
and practical (\ie, not requiring access to the model's weights). Finally, we show that it can even align image and
text features computed from uni-modal models.
\end{itemize}
\vspace{-0.5cm}

\begin{table}[h]
    \caption{\textbf{Training Time:} train time for CoOp \cite{zhou2022learning} and for the proposed LFA on
    ImageNet (16-shot) using ViT-B/16 as the visual encoder on a single V100 GPU.}
    \label{tab:train-time}
    \centering
    \resizebox{1.0\linewidth}{!}{%
    \begin{tabular}{lll}
    \toprule
    Method & Training Time & Test Acc. \\ 
    \midrule
    CoOp & 3h 22min & 71.92 \\
    \midrule
    LFA (Feature Extraction) \hspace{0.8in} & 2min 37s \\ 
    LFA (Procrustes Initialisation) & 4s \\
    LFA (Refinement) & 28s \\
    \rowcolor{tabhighlight}  LFA (Total) & \textbf{3min 9s} & \textbf{72.61} \\
    \bottomrule
    \end{tabular}
    }
\end{table}

\section{Related Work}

\noindent\textbf{Vision-Language (V-L) Models:} Recently, we have witnessed
an explosion of research in V-L foundation models, including CLIP \cite{radford2021learning},
ALIGN \cite{jia2021scaling}, Florence \cite{yuan2021florence},
LiT \cite{zhai2022lit}, BASIC \cite{pham2021combined}, ALBEF \cite{li2021align} and CoCa \cite{yu2022coca}.
Such models are pre-trained on large amounts of image and text data
to learn a joint multi-modal embedding space.
After pre-training, they can be used on various downstream tasks in a few- or zero-shot setting.
For our work, we used CLIP~\cite{radford2021learning} to extract the frozen image and text features.

\vspace{0.05in}
\noindent\textbf{Learnable Prompts for V-L Models:}
Despite pre-training, V-L models still suffer from a modality gap~\cite{liang2022mind}
which is further exacerbated during downstream adaptation. To address this issue, recently,
soft prompt learning methods \cite{zhou2022learning,zhou2022conditional,bulat2022language,
zhu2022prompt,chen2022prompt,shu2022test,lu2022prompt,huang2022unsupervised} 
optimize a new set of learnable (soft) prompts to reduce the gap and align the visual and
text modalities. CoOp~\cite{zhou2022learning} was the first method to apply prompt
learning methods \cite{li2021prefix,lester2021power,han2022ptr,shin2020autoprompt}
popularized in NLP to V-L models. Subsequent works have improved upon this by incorporating
image conditioning for better generalization \cite{zhou2022conditional},
test-time adaptation \cite{shu2022test}, gradient matching \cite{shu2022test}, or by using
an additional text-to-text loss \cite{bulat2022language}. In contrast to all the aforementioned
methods, we propose to bridge the domain gap for a given downstream task directly in the feature space,
without requiring access to the model's weights nor expensive training procedures.

\vspace{0.05in}
\noindent\textbf{Linear Alignment:}
The problem of linearly aligning two sets of embeddings or high-dimensional real
vectors is a well-studied problem in machine learning, with various applications
in computer vision and NLP. Classical applications range from sentence
classification \cite{fung1995compiling,rapp1995identifying},
to shape and motion extraction \cite{tomasi1992shape}, registration \cite{cootes1995active}
and geometrical alignment \cite{fischler1981random,leordeanu2005spectral,liu2008sift}.
In vision, linear mappings are widely used for zero-shot learning
\cite{frome2013devise,akata2013label,akata2015evaluation,romera2015embarrassingly}
for aligning the image features and their class attributes.
In NLP, and after the introduction of word embeddings \cite{mikolov2013efficient,bojanowski2017enriching},
this linear alignment problem was revisited and extensively studied, and improved upon for the task
of word translation \cite{artetxe2016learning,grave2019unsupervised,hoshen2018non,conneau2017word,
xing2015normalized,mikolov2013exploiting,cao2016distribution,joulin2018loss,zhang2017earth,
smith2017offline,artetxe2018acl,artetxe2018aaai,artetxe2017acl,artetxe2016emnlp}.
In this paper, we take strong inspiration from this line of work and set to
adapt them for the case of V-L models.

\section{Motivation: Approximating Soft Prompts with a Linear Transformation}
\label{sec:approximating_prompting}

Herein, we empirically show that the V-L alignment effect achieved by prompt learning can
be approximated by finding a simple linear transformation
$\mathbf{W} \in \mathbb{R}^{d \times d}$
that maps the class prototypes computed from the class names only (\ie, the class name text embeddings) 
to the ones obtained by soft prompt learning. To demonstrate this, let $\mathbf{Y} \in \mathbb{R}^{C \times d}$ be the class name embeddings
represented in matrix form,
and similarly, let $\mathbf{Y}^\prime \in \mathbb{R}^{C \times d}$ be the class prototypes
obtained by soft prompt learning. Our objective is to learn a linear transformation $\mathbf{W} \in \mathbb{R}^{d \times d}$
that tries to approximate prompt
learning, \ie, $\mathbf{Y} \xrightarrow{\mathbf{W}} \mathbf{Y}^\prime$,
by solving the following least square problem:
\vspace{-0.2cm}
\begin{equation}
\label{eq:w-prompt}
\min_{\mathbf{W} \in \mathbb{R}^{d \times d}}\| \mathbf{Y} \mathbf{W} - \mathbf{Y}^\prime \|_\mathrm{F}^2,
\end{equation}
where $\|\cdot\|_\mathrm{F}$ is the Frobenius norm.

If the classification results are consistent when using either $\mathbf{Y} \mathbf{W}$
or $\mathbf{Y}^\prime$ as class prototypes, then we could use $\mathbf{Y} \mathbf{W}$ to approximate the V-L realignment
achieved by prompt optimization. As shown in \cref{fig:intro} (1b) and (1c), $\mathbf{W}$ can almost perfectly
approximate the effects of prompt learning resulting in the same test set accuracy (\ie, same accuracy on DTD and
39.9 vs. 40.1 on FGVC Aircraft).

Note that, in practice, we want to avoid the training of the soft prompts. To this end, we can simply attempt to
learn a linear transformation directly from the image features to the class name text embeddings. This is the main
idea behind the proposed Linear Feature Alignment (LFA)
which directly finds a linear transformation for image-text alignment.

\section{Linear Feature Alignment}
\label{sec:alignment}

Our objective is to learn a linear mapping $\mathbf{W}$
for aligning the image embeddings $\mathbf{X}$ with their corresponding text class prototypes $\mathbf{Y}$,
\ie, $\mathbf{X} \xrightarrow{\mathbf{W}} \mathbf{Y}$. 
Once $\mathbf{W}$ is learned, in order to classify a new sample $\mathbf{x}$, we
obtain its $C$-way class probabilities from
$\operatorname{softmax}(\mathbf{x}\mathbf{W} \cdot \mathbf{Y}^\top / \tau)$ with $\tau$
being the temperature parameter. To learn $\mathbf{W}$, LFA firstly uses for initialization a closed-form solution to
a least-squares optimization problem, then minimizes a re-ranking loss to refine the initial solution.
LFA is described in detail in the following sections.

\subsection{Problem Formulation}

Let $\mathbf{X} \in \mathbb{R}^{N \times d}$ be the image embeddings of $N$ examples
produced by the CLIP image encoder, and let $\mathbf{Y} \in \mathbb{R}^{C \times d}$ be the $C$ class prototypes corresponding
to the encoded class names using CLIP text encoder (\ie, without any prompts). Moreover, let $\mathbf{P} \in \mathcal{P}_{N \times C}$ be
an assignment matrix that assigns each class prototype to its corresponding image embedding with
$\mathcal{P}_{N \times C} =\{\mathbf{P} \in\{0,1\}^{N \times C},\ 
\mathbf{P} \mathbf{1}_C = \mathbf{1}_N\}$
as the set of binary permutation matrices that map each one of the $N$ rows to one of the $C$ columns,
\ie, the input images to their corresponding classes. In a supervised setting where we are provided with $N$ (image-class) pairs,
$\mathbf{P}$ is the stacked $N$ $C$-dimensional one-hot vectors.

Our objective is to find an optimal linear mapping that bridges the modality gap and aligns each image embedding with its text class prototype.
To this end, the linear mapping can be learned by solving the following least squares:
\begin{equation}
\label{eq:main-obj}
\underset{\mathbf{W} \in \mathbb{R}^{d \times d}}{\operatorname{argmin}}
\|\mathbf{X} \mathbf{W} - \mathbf{P} \mathbf{Y} \|_\mathrm{F}^2.
\end{equation}
This is the standard Procrustes analysis formulation that aims to find
a linear transformation between two sets of points $\mathbf{X}$ and $\mathbf{PY}$.

\subsection{Orthogonal Procrustes}

It is common to impose
further constraints on the mapping
$\mathbf{W}$ to adapt it to the task at hand.
Of particular interest is the orthogonality constraint that
has been shown empirically to be well-suited for mappings between different
word embeddings and to result in improved alignment~\cite{xing2015normalized}.
By enforcing the orthogonality constraint on $\mathbf{W}$, \cref{eq:main-obj}
becomes an Orthogonal Procrustes (OP) analysis optimization problem:
\begin{equation}
\label{eq:orth-proc}
\mathbf{W}_{\mathrm{op}} = \underset{\mathbf{W} \in \mathcal{O}_d}{\operatorname{argmin}}
\|\mathbf{X}\mathbf{W} - \mathbf{P} \mathbf{Y} \|_\mathrm{F}^2,
\end{equation}
with $\mathcal{O}_d = \{\mathbf{W} \in \mathbb{R}^{d \times d},\ 
\mathbf{W}^\top \mathbf{W} = \mathbf{I}_d \}$
as the set of orthogonal matrices and $\mathbf{I}_d$
as the $d$-dimensional identity matrix.
As shown in \cite{schonemann1966generalized}, under this constraint, 
\cref{eq:orth-proc} admits a closed-form solution from the singular value
decomposition (SVD) of $\mathbf{X}^\top \mathbf{P} \mathbf{Y}$:
\begin{equation}
\begin{aligned}
\label{eq:orth-proc-solution}
\mathbf{W}_{\mathrm{op}} &= \underset{\mathbf{W} \in \mathcal{O}_d}{\operatorname{argmin}}
\|\mathbf{X} \mathbf{W} - \mathbf{P} \mathbf{Y} \|_\mathrm{F}^2
= \mathbf{U} \mathbf{V}^\top, \\
&\text{with}\ \operatorname{SVD}(\mathbf{X}^\top\mathbf{P}\mathbf{Y}) =
\mathbf{U \Sigma} \mathbf{V}^\top.
\end{aligned}
\end{equation}
Moreover, under the orthogonality constraint,
the obtained mapping preserves the vector dot product and their $\ell_2$ distances, thus making it
suitable for V-L models trained with a contrastive loss.

\subsection{$\beta$-Procrustes}

\begin{figure}[t]
  \centering
  \includegraphics[width=0.45\textwidth]{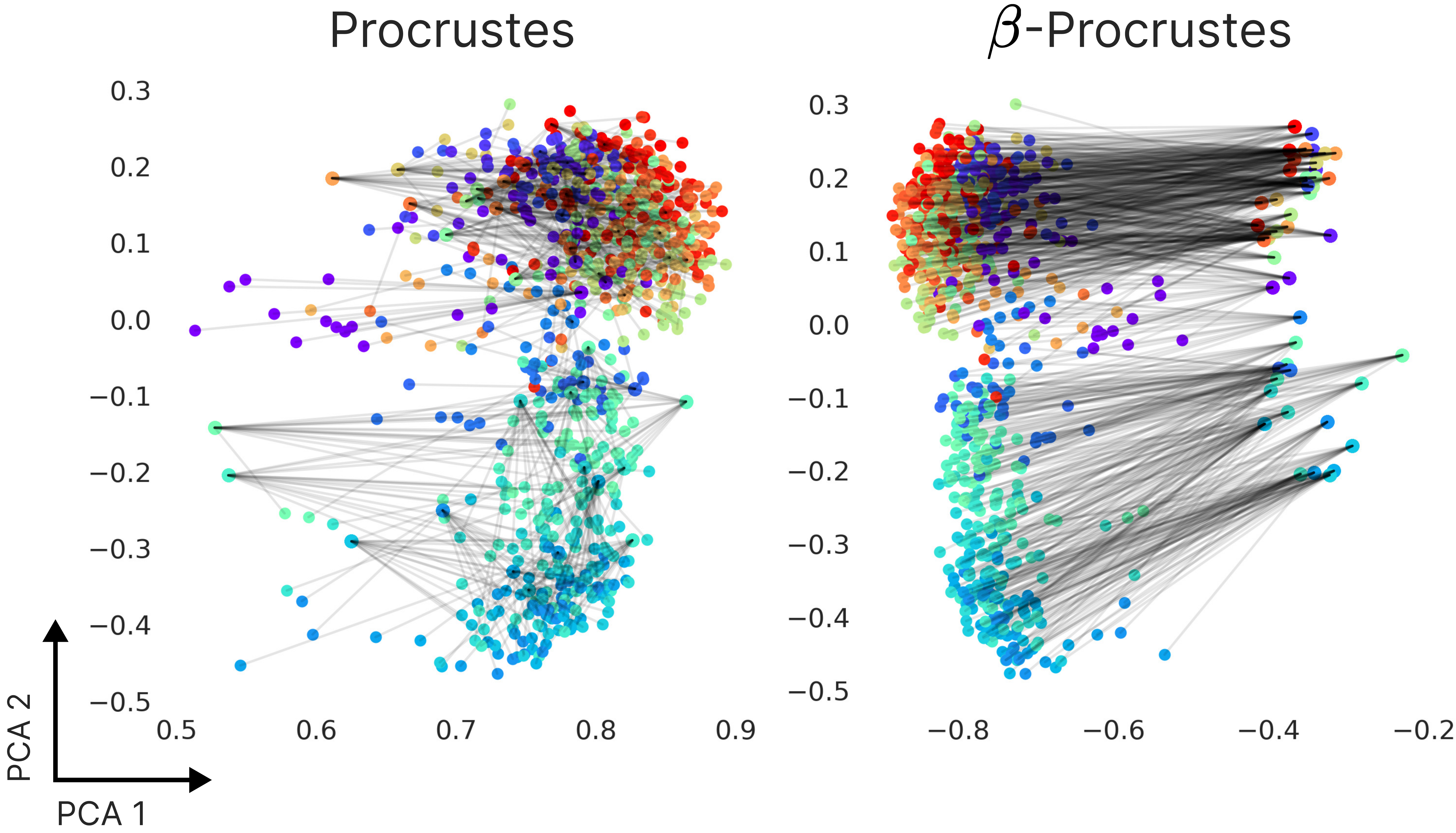}
  \vspace{-0.051in}
  \caption{\textbf{Effect of $\beta$-Procrustes:} By pushing the orthogonal
  Procrustes solution of \cref{eq:orth-proc-solution} towards an identity mapping
  via the update rule in \cref{eq:beta-procrustes}, we avoid 
  the overfitting exhibited using the original solution and obtain
  better alignment (\ie as suggested by the observed class prototypes-image embeddings cross-interference).
  Here, each class prototype and its image embeddings share
  the same color and the embeddings shown are from 50 randomly sampled ImageNet classes.}
  \vspace{-0.05in}
  \label{fig:beta-procrustes}
\end{figure}

\begin{table}[t]
  \caption{\textbf{$\beta$-Procrustes:} Top-1 acc. for
  16-shot per class to obtain $\mathbf{W}$.}
  \vspace{-0.1in}
  \label{tab:ablation-beta-procrustes}
  \vspace{5pt}
  \centering
  \resizebox{\linewidth}{!}{%
  \begin{tabular}{l*{5}c}
  \toprule
  Method & ImageNet & Aircraft & DTD & Food101 & Caltech101 \\ 
  \midrule
  CLIP ViT-B/16  & 62.8 & 22.1 & 45.1 & 83.9 & 88.0 \\
  Procrustes & 52.5 & 27.9 & 63.4 & 78.8 & 93.4 \\
  $\beta$-Procrustes & \textbf{64.8} & \textbf{29.1} & \textbf{65.8} & \textbf{85.5} & \textbf{94.7} \\
  \bottomrule
  \vspace{-0.25in}
  \end{tabular}
}
\end{table}

The orthogonal Procrustes solution is efficient
and easy to compute, however, it suffers from extreme overfitting, especially if the initial modality gap
is small. On ImageNet, for instance, and as shown in
\cref{fig:beta-procrustes} and \cref{tab:ablation-beta-procrustes}, 
the orthogonal Procrustes solution results in overly entangled class prototypes
and a lower test accuracy than the original CLIP features (\ie, 62.8 $\rightarrow$ 52.5).
To solve this, we propose $\beta$-Procrustes, a regularized Procrustes solution
that is pushed to be close to an identity mapping via the following update:
\begin{equation}
\label{eq:beta-procrustes}
\mathbf{W}_{\beta} \leftarrow \mathbf{W}_{\mathrm{op}} - \beta (\mathbf{W}_{\mathrm{op}}- \mathbf{I}_d),
\end{equation}
where $\beta \in [0,1]$ is an interpolation hyperparameter between an 
identity mapping ($\beta=1$) and the orthogonal solution of \cref{eq:orth-proc-solution} ($\beta=0$).
This update is equivalent to a single gradient descent step of the regularization term
$R_\beta(\mathbf{W}_\mathrm{op}) = 
\frac{\beta}{2} \|\mathbf{W}_\mathrm{op} - \mathbf{I}_d \|_\mathrm{F}^2$, \ie
$\nabla_{\mathbf{W}_\mathrm{op}} R_\beta(\mathbf{W}_\mathrm{op})=
\beta (\mathbf{W}_{\mathrm{op}}- \mathbf{I}_d)$.
As shown in \cref{fig:beta-procrustes} and \cref{tab:ablation-beta-procrustes},
this simple update results in better alignment and improved test accuracy. 
For the choice of the hyperparameter $\beta$, it can be found via cross-validation on the training set
or set to a fixed value with $\beta \in [0.6, 0.9]$ without a significant impact on the results.
The hyperparameter $\beta$ can be determined through cross-validation on the training set or set to a
fixed value $\beta \in [0.6, 0.9]$ without significantly impacting the results.

\subsection{Mapping Refinement}

\begin{figure}[t]
  \centering
  \includegraphics[width=0.48\textwidth]{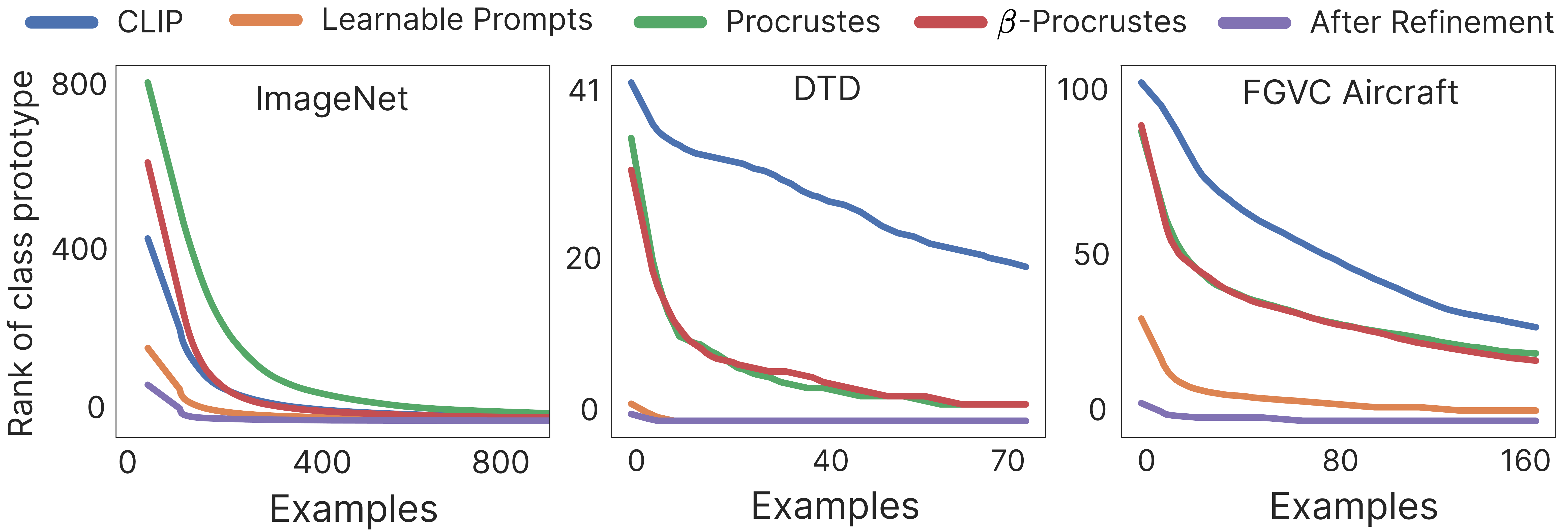}
  \caption{\textbf{Hubness:} We show the rank of the ground-truth class prototypes
  of different training examples. We see that even after the $\beta$-Procrustes alignment
  step, the embedding space still contains a high number of hubs. After a
  refinement step, we reduce the hubness and obtain better image-class prototype alignment
  than soft-prompts.}
  \vspace{-0.1in}
  \label{fig:hubness}
\end{figure}

While $\beta$-Procrustes improves the results, they are still
not on par with those obtained with the soft prompts. To investigate why,
in \cref{fig:hubness}, we show the rank of the ground-truth class prototype
for different training examples, and we observe that even
after the alignment with $\beta$-Procrustes, many examples
have high ranks, \ie, they are closer to many other class prototypes
than their own. At inference, this will result in many misclassifications.
This is a known problem in many retrieval cases \cite{aucouturier2008scale,jegou2008accurate},
and is often caused by the \textit{hubness problem}
\cite{radovanovic2010hubs,dinu2014improving,conneau2017word,joulin2018loss}.
Hubs are points (\eg, class prototypes)
in the high dimensional vector space that are the nearest neighbors of many other points
(\eg, image embeddings), and as a result, they greatly influence the classification
probabilities (and thus the accuracy at test time).

To mitigate this effect, and inspired by popular metric learning losses 
\cite{hadsell2006dimensionality,sun2020circle,wang2018cosface,wang2017deep,cakir2019deep},
we propose to refine the mapping $\mathbf{W}$ 
by optimizing an Adaptive Reranking (ARerank) loss designed specifically
to reduce the hubness problem, and defined as follows:
\begin{equation}
\label{eq:adaptive-rerank}
\begin{aligned}
\mathcal{L}_{\mathrm{ARerank}} (\mathbf{x}_i \mathbf{W}, \mathbf{Y}) &=
  \frac{1}{k} \sum_{\mathbf{y}_j \in \mathcal{N}_k(\mathbf{x}_i\mathbf{W})} \ell_{ij} \\
\text{where} \ \ \ell_{ij} &= \max \{d_{ii} - d_{ij} + m_{ij}, 0\},
\end{aligned}
\end{equation}
where $d_{ii} = \|\mathbf{x}_i \mathbf{W} -\mathbf{y}_{c_i}\|_2$ is the $\ell_2$ distance
between the aligned image embedding $\mathbf{x}_i \mathbf{W}$ and its class prototype $\mathbf{y}_{c_i}$,
and similarly, $d_{ij} = \|\mathbf{x}_i \mathbf{W} -\mathbf{y}_j\|_2$
 the $\ell_2$ distance
between the aligned image embedding and each of its $k$ nearest class prototypes
$\mathbf{y}_j \in \mathcal{N}_k(\mathbf{x}_i\mathbf{W})$, and $m_{ij}$ the margin.
Empirically, we found that $k = 3$ works well for most datasets.

To make the re-ranking dynamic and
avoid having multiple hyperparameters, we opt for an adaptive margin selection
approach similar to \cite{li2020boosting}. Specifically, the margin
between image $i$ and a given class prototype $j$ is
defined based on the cosine similarity between its class prototype
$\mathbf{y}_{c_i}$ and the $j$-th class prototype, \ie,
$m_{ij} = (1.0 - \mathbf{y}_{c_i}^\top \mathbf{y}_j) / s$,
where $s$ is a scalar set to 4 for all experiments. 
By doing so, we ensure that each image embedding is pushed away from nearby incorrect class
prototypes with an adaptive margin, while the distance to its class prototype is kept unchanged,
thus mitigating the hubness problem and avoiding learning a degenerate mapping. 
As shown in \cref{tab:ablation-refinement-loss}, ARerank loss
outperforms standard embedding optimization losses, and also demonstrates better results than the CSLS criterion proposed by
\cite{conneau2017word} used to reduce hubness for word translation. Finally, as shown in \cref{tab:ablation-procrustes-refine}, the coupling of $\beta$-Procrustes and ARerank-based refinement results in better
accuracy.

\begin{table}[t]
    \caption{\textbf{Refinement Loss:} Top-1 acc. for
  16-shot per class.}
    \vspace{-0.15in}
    \label{tab:ablation-refinement-loss}
    \vspace{5pt}
    \centering
    \resizebox{\linewidth}{!}{%
    \begin{tabular}{l*{5}c}
    \toprule
    Refinement Loss & ImageNet & Aircraft & DTD & Food101 & Caltech101 \\ 
    \midrule
    {CLIP ViT-B/16} & 62.8 & 22.1 & 45.1 & 83.9 & 88.0 \\
    {$\beta$-Procrustes} & {64.8} & {29.1} & {65.8} & {85.5} & {94.7} \vspace{0.05in}\\
    Contrastive & 65.5 & 40.1 & 71.5 & 85.7 & 92.2 \\
    Triplet with margin & 70.7 & 43.5 & 72.6 & 86.9 & \textbf{95.9} \\
    CSLS & 71.1 & 40.9 & 72.2 & 86.8 & \textbf{95.9} \\
    ARerank & \textbf{71.7} & \textbf{45.1} & \textbf{72.7} & \textbf{87.5} & \textbf{95.9} \\
    \bottomrule
    \end{tabular}
    }
\end{table}

\begin{table}[t]
    \caption{\textbf{$\beta$-Procrustes \& Mapping Refinement:} Top-1 acc. for
    16-shot per class.}
    \vspace{-0.15in}
    \label{tab:ablation-procrustes-refine}
    \vspace{5pt}
    \centering
    \resizebox{\linewidth}{!}{%
    \begin{tabular}{l*{5}c}
    \toprule
    Method & ImageNet & Aircraft & DTD & Food101 & Caltech101 \\ 
    \midrule
    CLIP  $\rightarrow$ Refine  & 71.6 & 43.4 & 72.6 & 87.2 & \textbf{95.9} \\
    CLIP  $\rightarrow$ Proc. $\rightarrow$ Refine & 70.7 & 44.8 & 72.4 & 85.3 & 95.7 \\
    CLIP  $\rightarrow$ $\beta$-Proc. $\rightarrow$ Refine & \textbf{71.7} & \textbf{45.1} & \textbf{72.7} & \textbf{87.5} & \textbf{95.9} \\
    \bottomrule
    \end{tabular}
    }
\end{table}

\begin{algorithm}[t]
  \small
  \caption{\small Linear Feature Alignment (LFA) }
  \label{alg:LFA}
  \lstset{
    backgroundcolor=\color{white},
    basicstyle=\fontsize{6.5pt}{6.5pt}\ttfamily\selectfont,
    columns=fullflexible,
    breaklines=true,
    captionpos=b,
    commentstyle=\fontsize{7.5pt}{7.5pt}\color{DarkIndigoPurple},
    keywordstyle=\fontsize{7.5pt}{7.5pt}\color{indianred},
    escapechar={|}, 
  }
  \begin{lstlisting}[language=python]
  def LFA(img_feats, cls_prototypes, labels, beta, test_img_features):
    """
    img_feats: [N, d]
    cls_prototypes: [C, d]
    labels: [N]
    test_img_features: [M, d]
  
    # N: number of training image features
    # C: number of classes
    # d: features dimensionality
    # M: number of test image features
    """
  
    # One-to-one matchings
    text_feats = cls_prototypes[labels]
    
    # Orthogonal Procrustes
    u, _, v = torch.svd(img_feats.T @ text_feats)
    W_op = u @ v.T
  
    # Beta-Procrustes
    identity = torch.eye(d)
    W_beta = W_op - (W_op - identity) * beta
  
    # Refine
    W = adaptive_rerank_refine(W_beta)
  
    test_logits = (test_img_features @ W) @  cls_prototypes.T
    test_preds = test_logits.argmax(-1)
    
    return test_preds
\end{lstlisting}
\end{algorithm}

\subsection{Overall LFA algorithm}\label{ssec:overall_lfa}

Herein, we define the overall algorithm obtained by combining the steps defined in the previous sections.
We consider two cases: supervised learning, where labeled data is provided, and unsupervised
learning, where only unlabeled images are available.

\vspace{0.05in}
\noindent \textbf{Supervised Alignment:}
\label{sec:sup-alignment} In a supervised setting, we can directly construct the
assignment matrix $\mathbf{P}$
between the image embeddings $\mathbf{X}$ and class prototypes $\mathbf{Y}$
using the ground-truth data. The overall algorithm can be then defined as (see also Alg.~\ref{alg:LFA}):
\begin{enumerate}
\itemsep0em
\item $\mathbf{W}_{\mathrm{op}} \leftarrow$ Orthogonal Procrustes [$\mathbf{X}, \mathbf{P}\mathbf{Y}$]
(\cref{eq:orth-proc-solution}).
\item $\mathbf{W}_\beta \leftarrow$ $\beta$-Procrustes[$\mathbf{W}_{\mathrm{op}}$]
(\cref{eq:beta-procrustes}).
\item $\mathbf{W} \leftarrow$ Refine [$\mathbf{W}_\beta$, $\mathbf{X}, \mathbf{P}\mathbf{Y}$]
(\cref{eq:adaptive-rerank}).
\end{enumerate}

\noindent \textbf{Unsupervised Alignment:}\label{sec:unsup-alignment} 
In an unsupervised setting, the correspondences between the 
image embeddings and their class prototypes are not known a priori. Thus
the assignment matrix $\mathbf{P}$ must be estimated jointly with learning the mapping $\mathbf{W}$.
By keeping the orthogonality constraint over $\mathbf{W}$ and solving for
both $\mathbf{P}$ and $\mathbf{W}$, the resulting optimization problem,
often referred to as the Wasserstein-Procrustes
\cite{zhang2017earth,grave2019unsupervised} problem takes the following form:
\begin{equation}
\label{eq:wasser-proc}
\mathbf{W}^\star, \mathbf{P}^\star =
\underset{\mathbf{W} \in \mathcal{O}_d, \mathbf{P} \in \mathcal{P}_{N \times C}}{\operatorname{argmin}}
\|\mathbf{X}\mathbf{W} - \mathbf{P} \mathbf{Y} \|_\mathrm{F}^2.
\end{equation}
As neither of the two sets $\mathcal{O}_d$
and $\mathcal{P}_{N \times C}$ are convex, this optimization problem is not
convex either. To solve it, in practice, we follow a simple heuristic by alternating between
finding the assignments $\mathbf{P}$ using the efficient Sinkhorn algorithm~\cite{cuturi2013sinkhorn} and 
refining the mapping following \cref{eq:adaptive-rerank}. Given this,
Unsupervised LFA (U-LFA) takes the following form:

\begin{enumerate}
\vspace{-0.05in}
\itemsep0em
\item $\mathbf{P} \leftarrow$ Sinkhorn [$\mathbf{X}, \mathbf{Y}$].
\item $\mathbf{W}_{\mathrm{op}} \leftarrow$ Orthogonal Procrustes [$\mathbf{X}, \mathbf{P}\mathbf{Y}$].
\item $\mathbf{W} \leftarrow$ $\beta$-Procrustes[$\mathbf{W}_{\mathrm{op}}$].
\item Repeat for $n$ iterations:
  \begin{enumerate}
    \itemsep0em
    \vspace{-0.05in}
    \item $\mathbf{P} \leftarrow$ Sinkhorn [$\mathbf{X}\mathbf{W}, \mathbf{Y}$].
    \item $\mathbf{W} \leftarrow$ Refine [$\mathbf{W}$, $\mathbf{X}, \mathbf{P}\mathbf{Y}$].
  \end{enumerate}
\end{enumerate}

\begin{table}[t]
    \caption{\textbf{U-LFA:} Top-1 acc. for
    16-shot per class (without labels):}
    \vspace{-0.15in}
    \label{tab:unsup-alignment}
    \vspace{5pt}
    \centering
    \resizebox{\linewidth}{!}{%
    \begin{tabular}{l*{5}c}
    \toprule
    Refinement Loss & ImageNet & Aircraft & DTD & Food101 & Caltech101 \\ 
    \midrule
    {CLIP ViT-B/16} & 62.8 & 22.1 & 45.1 & 83.9 & 88.0 \\
    U-LFA ($n=1$) & 66.9 & \textbf{24.5} & \textbf{48.0} & \textbf{86.7} & 94.4 \\
    U-LFA ($n=5$) & \textbf{68.6} & 24.2 & 47.5 & 86.1 & \textbf{95.1} \\
    \midrule
    \multicolumn{6}{l}{\textit{with template: a photo of a \{{cls name}\}.}} \\
    {CLIP ViT-B/16} & 66.8 & 23.3 & 43.9 & 85.8 & 92.9  \\
    U-LFA ($n=1$) & 69.1 & 27.9 & \textbf{50.2} & \textbf{87.4} & 95.3 \\
    U-LFA ($n=5$) & \textbf{70.3} & \textbf{28.0} & 49.0 & 86.6 & \textbf{95.5} \\
    \bottomrule
    \end{tabular}
    }
\end{table}

\begin{table}[t]
    \caption{\textbf{LFA analysis under distribution shift:} Top-1 acc. for 16 shots
    per class on the Base (B) and New (N) sets and two IN variants. IN refers to ImageNet.}
    \vspace{-0.15in}
    \label{tab:adaptation}
    \vspace{5pt}
    \centering
    \resizebox{\linewidth}{!}{%
    \begin{tabular}{l*{9}c}
    \toprule
    \multirow{3}*{Method} &
    \multicolumn{6}{c}{Language Shift} & & \multicolumn{2}{c}{Image Shift} \\
    \cmidrule{2-7} \cmidrule{9-10}
    & \multicolumn{2}{c}{ImageNet} &
    \multicolumn{2}{c}{Aircraft} &
    \multicolumn{2}{c}{DTD} &
    \multicolumn{2}{c}{} \\
    & B & N & B & N & B & N & & IN-A & IN-R \\ 
    \midrule
    LFA & \textbf{76.9} & 67.1 & \textbf{41.5} & 30.7 & \textbf{81.6} & 54.9 && 49.7 & 74.5 \\
    LFA (average) & 76.8 & 68.6 & 40.7 & 31.8 & \textbf{81.6} & 59.7 &&  50.6 & 75.5 \\
    LFA (two mappings) & \textbf{76.9} & \textbf{69.4} & \textbf{41.5} & \textbf{32.3} & \textbf{81.6} & \textbf{60.6} && \textbf{51.5} & \textbf{76.1} \\
    \bottomrule
    \end{tabular}
}
\end{table}

\subsection{LFA for Base-to-New (Zero-Shot) Recognition}
\label{subsec:two-mappings}

\begin{table}[t]
    \caption{\textbf{Augmentations:} average Top-1 acc. for 16-shot per class
    on 11 datasets using 5 crops per image, with and without
    a prompt template.}
    \vspace{-0.15in}
    \label{tab:ablation-augmentation}
    \vspace{5pt}
    \centering
    \resizebox{\linewidth}{!}{%
    \begin{tabular}{l*{4}c}
    \toprule
    Backbone & RN50 & RN101 & ViT-B/32 & ViT-B/16 \\ 
    \midrule
    LFA  & 74.20 & 76.83 & 76.69 & 80.54 \\
    LFA + 5 crops & 74.49 & 77.05 & 76.95 & 80.88 \\
    LFA + 5 crops + template & \textbf{74.75} & \textbf{77.14} & \textbf{77.17} & \textbf{81.21}\\
    \bottomrule
    \vspace{-0.3in}
    \end{tabular}
    }
\end{table}

\begin{table*}[ht]
\caption{\textbf{Few-shot Classification:} Top-1 acc. for
    16-shot per class when using CLIP, CoOp \cite{zhou2022learning} and LFA, with either RN50, RN101, ViT-B/32 or ViT-B/16 as the vision encoder.}
\label{tab:backbones}
\vspace{-0.1in}
\centering
\resizebox{\linewidth}{!}{%
\begin{tabular}{l*{12}cc}
\toprule
Method & Pets & Flowers102 & Aircraft & DTD & EuroSAT & Cars & Food101 & SUN397 & Caltech101 & UCF101 & ImageNet & Avg. & $\Delta$ \\ 
\midrule
CLIP RN50 & 85.77 & 66.14 & 17.28 & 42.32 & 37.56 & 55.61 & 77.31 & 58.52 & 86.29 & 61.46 & 58.18 & 58.77 \\
CoOp & 86.16 & {\bf94.80} & 32.29 &	63.16 &	83.55 &	73.27 &	74.46 &	69.12 &	91.62 &	75.29 &	63.08 &	73.35 \\
\rowcolor{tabhighlight}
\textbf{LFA} & {\bf86.75} & 94.56 & {\bf35.86} & {\bf66.35} & {\bf84.13} & {\bf73.58} & {\bf76.32} & {\bf71.32} & {\bf92.68} & {\bf77.00} & {\bf63.65} & {\bf74.75} & {\bf\textcolor{IndigoPurple}{+1.40}} \\
\midrule
CLIP RN101 & 86.75	& 64.03	& 18.42 &	38.59 &	32.59	 & 66.23 &	80.53 &	58.96	& 89.78 &	60.96	& 61.62	& 59.86 \\
CoOp & 88.57	& {\bf95.19}	& 34.76 &	65.47 &	{\bf83.54}	 & {\bf79.74} &	79.08 &	71.19	& 93.42 &	77.95	& 66.60	& 75.96 \\
\rowcolor{tabhighlight} 
\textbf{LFA}  & {\bf88.80} & 93.11 & {\bf39.62} & {\bf68.95} & 83.43 & 79.45 & {\bf81.57} & {\bf72.69} & {\bf94.53} & {\bf79.28} & {\bf67.16} & {\bf77.14} & {\bf\textcolor{IndigoPurple}{+1.18}} \\
\midrule
CLIP ViT-B/32 & 87.49	& 66.95	& 19.23 &	43.97 &	45.19	 & 60.55 &	80.50 &	61.91	& 90.87 &	62.01	& 62.05	& 61.88 \\
CoOp & {\bf88.68}	& {\bf94.97}	& 33.22 &	65.37 &	83.43	 & 76.08 &	78.45 &	72.38	& 94.62 &	78.66	& 66.85	& 75.70 \\
\rowcolor{tabhighlight} 
\textbf{LFA}  & 88.62 & 93.84 & {\bf38.01} & {\bf68.87} & {\bf83.88} & {\bf76.72} & {\bf81.31} & {\bf74.12} & {\bf95.10} & {\bf80.81} & {\bf67.63} & {\bf77.17} & {\bf\textcolor{IndigoPurple}{+1.47}} \\
\midrule
CLIP ViT-B/16 & 89.21	& 71.34	& 24.72 &	44.39 &	47.60	 & 65.32 &	86.06 &	62.50	& 92.94 &	66.75	& 66.73	& 65.23 \\
CoOp & {\bf92.53} & 96.47	& 42.91 &	68.50 &	80.87	 & {\bf83.09} &	{\bf87.21} &	75.29	& 95.77 &	82.24	& 71.92	& 79.71 \\
\rowcolor{tabhighlight} 
\textbf{LFA}  & 92.41 & {\bf96.82} & {\bf46.01} & {\bf71.89} & {\bf87.31} & 82.23 & 87.14 & {\bf76.65} & {\bf96.24} & {\bf83.99} & {\bf72.61} & {\bf81.21} & {\bf\textcolor{IndigoPurple}{+1.50}} \\
\bottomrule
\end{tabular}
}
\end{table*}

\begin{table*}[t]
\caption{\textbf{Out-of-Domain Generalization:} the obtained average Top-1 acc. on various ImageNet variants
after training on ImageNet (\ie, source) using 16-shot training data per class and using RN50 as the visual encoder.
We show the results for the baseline CLIP \cite{radford2021learning}, CoOp \cite{zhou2022learning},
CoCoOp \cite{zhou2022conditional}, and the proposed LFA on each dataset, their average and the
Out-of-Distribution (OOD) average (\ie, over the target datasets).}
\label{tab:img-ood}
\vspace{-0.1in}
\centering
\resizebox{0.9\linewidth}{!}{%
\begin{tabular}{l*{8}cc}
\toprule
& Source && \multicolumn{4}{c}{Target} & & \\
\cmidrule{2-2} \cmidrule{4-7}
Method      & ImageNet   &&  ImageNet-A  &  ImageNet-V2.  & ImageNet-R. & ImageNet-Sketch   & Avg.  & OOD Avg. & $\Delta$ \\
\midrule
CLIP-RN50       &  58.16  &&   21.83      &     51.41        & 56.15    &      33.37    &  44.18   &    40.69      \\
CoOp            &  63.33 &&  23.06  &   55.40     &  56.60   &  \textbf{34.67}   &     46.61   &     42.43       \\
CoCoOp          &  62.81 &&  23.32  &   55.72     &  57.74  &  34.48   &     46.81   &    42.82   \\
\rowcolor{tabhighlight} 
\textbf{LFA}    &  \textbf{63.88} && \textbf{24.31}  & \textbf{55.79} & \textbf{58.13}  &  34.37 & \textbf{47.29} &  \textbf{43.15}  & \textcolor{IndigoPurple}{\textbf{+0.32}} \\
\midrule
CLIP-ViT-B/16   &  66.73 &&  47.87  &   60.86     &  73.98   &  46.09   &   59.11     &   57.2  \\
CoOp            &  71.51 &&  49.71  &   64.20     &  75.21   &  47.99   &   61.72     &   59.28  \\
CoCoOp          &  71.02  && 50.63  & 64.07       &  \textbf{76.18}   &  \textbf{48.75}   &   62.13     &   59.91  \\
\rowcolor{tabhighlight} 
\textbf{LFA}    & \textbf{72.65} && \textbf{51.50} & \textbf{64.72} & 76.09 & 48.01 & \textbf{62.59} &  \textbf{60.08} & \textcolor{IndigoPurple}{\textbf{+0.17}} \\
\bottomrule
\end{tabular}
}
\end{table*}

\begin{table}[t]
\caption{\textbf{Few-shot Action Recognition:} Top-1 acc. for 8 and 16-shot per class. We compare against our implementation of Video Prompting \cite{ju2022prompting}, which
is trained either with mean over the per-frame features (\ie, Temporal: \xmark)
or a single Transformer layer (\ie, Temporal: \checkmark).}
\label{tab:action-recognition-fewshot}
\vspace{-0.1pt}
\centering
\resizebox{\linewidth}{!}{%
\begin{tabular}[t]{llccccccc}
\toprule
N-shot & Method & Soft-Prompt & Temporal & UCF-101 & HMDB51 & Avg. & $\Delta$\\
\midrule
& CLIP~\cite{radford2021learning,ju2022prompting} & hand-craft & \xmark & 64.7 & 40.1 & 52.40  \\
\midrule
\multirow{3}*{8} &
Video Prompting & \checkmark & \xmark & 73.37 & 49.72 & 61.54 \\
& Video Prompting & \checkmark & \checkmark & 86.21 & 60.52 & 73.36 \\
\rowcolor{tabhighlight} 
& \textbf{LFA} & \xmark & \xmark & \textbf{87.60} & \textbf{60.74} & \textbf{74.17} & \textbf{\textcolor{IndigoPurple}{+0.81}} \\
\midrule
\multirow{3}*{16} &
Video Prompting & \checkmark & \xmark & 76.22 & 53.90 & 65.06 \\
& Video Prompting & \checkmark & \checkmark & \textbf{89.43} & \textbf{65.05} & \textbf{77.24} \\
\rowcolor{tabhighlight} 
& \textbf{LFA} & \xmark & \xmark & \textbf{89.47} & \textbf{65.08} & \textbf{77.27} & \textbf{+0.03} \\
\bottomrule
\end{tabular}
}
\vspace{-0.4cm}
\end{table}

\begin{table}[t]
\caption{\textbf{Action Recognition:} the obtained Top-1 acc. on the test sets of UCF101
and HMDB51 using the full training sets.
Video Prompting~\cite{ju2022prompting} is trained with two Transformer layers (\ie, Temporal: \checkmark)
on top of the frozen per-frame CLIP features to model the temporal information in the input video segment.
All results are obtained with ViT-B/16 as the visual encoder.}
\label{tab:action-recognition-full}
\vspace{-0.1pt}
\centering
\resizebox{1.0\linewidth}{!}{%
\begin{tabular}[t]{lcccccc}
\toprule
Method & Soft-Prompt & Temporal & UCF-101 & HMDB51 & Avg. & $\Delta$\\
\midrule
I3D~\cite{carreira2017quo} &&& 74.3 & 95.1 & 84.7 \\
S3D-G~\cite{Xie18-S3D} &&& 75.9 & 96.8 & 86.3 \\
R(2+1)D~\cite{Tran18} &&& 74.5 & 96.8 & 85.6 \\
R3D-50~\cite{hara2018can} &&& 66.0 & 92.0  & 79.0\\
\midrule
Video Prompting & \checkmark & \checkmark & 66.4 & \textbf{93.6} & 80.0 \\
\rowcolor{tabhighlight} 
\textbf{LFA} & \xmark & \xmark & \textbf{69.2} & 91.8 &  \textbf{80.5} & \textbf{\textcolor{IndigoPurple}{+0.5}} \\
\bottomrule
\end{tabular}
} 
\vspace{-0.6cm}
\end{table}

An important property of large-scale V-L models that recent few-shot adaptation methods seek to preserve is
their zero-shot generalization  ability, \ie, generalization from seen (base) classes to unseen (new) classes.
Training LFA in this setting, \ie, on the base set, may result in a mapping that fails to generalise to the new
set due to the distribution shift in between the two.

To address this, starting from a task-specific mapping $\mathbf{W}$, during the iterative refinement step,
we initialize a second $\mathbf{W}_\mathrm{tt}$ as an identity map $\mathbf{I}_d$. At each optimization step $t$ of
the refinement procedure, we then update $\mathbf{W}_\mathrm{tt}$
using $\mathbf{W}$ with an exponential moving average as follows:
\begin{equation}
\label{eq:tt-mapping}
\mathbf{W}_\mathrm{tt}(t) \leftarrow \alpha(t) \mathbf{W}_\mathrm{tt}(t) + (1 - \alpha(t)) \mathbf{W}(t)
\end{equation}
with $\alpha(t) \in [0, 1]$ as the momentum parameter, which is initialized as 0.9,
and is increased to 1.0 following a log schedule during the first half of the optimization.
This way, we only incorporate the first refinement updates into $\mathbf{W}_\mathrm{tt}$,
while the later ones, which tend to be more task-specific and may hinder generalization are largely ignored.
At test time, $\mathbf{W}$ can be used on the base classes, while $\mathbf{W}_\mathrm{tt}$
for the new classes. As shown in \cref{tab:adaptation}, this maintains the good accuracy on the base training domain,
while demonstrating good generalization when a distribution shift occurs (\ie, on novel classes).
Additionally, using a single mapping, obtained by taking the average
of $\mathbf{W}$ and $\mathbf{W}_\mathrm{tt}$ also achieves good results (see \cref{tab:adaptation}).

\section{Experiments}
\label{sec:experiments}

\noindent\textbf{Datasets \& Evaluation settings:}
For image classification, we consider the following evaluation protocols and settings: (1) standard few-shot classification,
as in~\cite{zhou2022learning}, (2) generalisation from base-to-new classes, where the model is trained in a
few-shot manner on the base classes and tested on a disjoint set of new classes, as in~\cite{zhou2022conditional},
and finally, (3) domain generalisation, where the model is trained on 
training set of ImageNet and is then tested on one of the four ImageNet variants with
some form of distribution shift, as in~\cite{zhou2022conditional,zhou2022learning}.
For standard few-shot
evaluation and generalisation from base-to-new classes, we report results on the 11 datasets used in CoOp~\cite{zhou2022learning}:
ImageNet~\cite{deng2009imagenet}, Caltech101~\cite{fei2004learning}, OxfordPets~\cite{parkhi2012cats},
StanfordCars~\cite{krause20133d}, Flowers102~\cite{nilsback2008automated}, Food101~\cite{bossard2014food},
FGVCAircraft~\cite{maji2013fine}, SUN397~\cite{xiao2010sun}, UCF101~\cite{soomro2012ucf101}, DTD~\cite{cimpoi2014describing} and
EuroSAT~\cite{helber2019eurosat}.
For domain generalisation, we follow previous work \cite{zhou2022learning,zhou2022conditional} and report the classification
results on four ImageNet variants: ImageNetV2 \cite{recht2019imagenet}, ImageNet-Sketch \cite{wang2019learning},
ImageNet-A \cite{hendrycks2021natural} and ImageNet-R \cite{hendrycks2021many}.

For action recognition, we align our setting with~\cite{ju2022prompting} and consider both standard (\ie using the full training set for
adaptation)
and few-shot classification settings 
and on two datasets, HMDB51~\cite{kuehne2011hmdb} and UCF101~\cite{soomro2012ucf101}\footnote{Both image classification
and action recognition experiments use UCF101, but take a single frame and a video segment respectively as input.}.
To get the video features to be aligned their class prototypes, we take the max aggregate on the per-frame CLIP features.

\vspace{0.05in}
\noindent\textbf{Implementation Details:} Unless stated otherwise, we base our experiments on a pre-trained
CLIP model~\cite{radford2021learning}. For each experiment, we pre-compute and save the image features alongside
the class prototypes and follow the adaptation procedure as described in \cref{ssec:overall_lfa}.
The class prototypes are formed by inserting the class name in the standard
templates~\cite{zhou2022learning} (\eg, ``\texttt{a photo of a \{{cls name}\}}''
for image tasks and ``\texttt{a video frame of a person \{{action type}\}}'' for videos). As it is common
practice to augment the images for prompting~\cite{zhou2022conditional,zhou2022learning}, for each training image,
we construct $c=5$ random cropped views, noting that a large $c$ is not crucial, as LFA still performs
well without them (\ie, $c=1$), as shown in \cref{tab:ablation-augmentation}.

\vspace{0.05in}
\noindent\textbf{Training Details:} 
For standard image few-shot experiments, we set $\beta$
based on cross-validation on the training set, while for the rest, we fix it to $\beta = 0.9$. For the refinement step,
we set $k=3$ for the ARerank loss, and finetune the mappings using AdamW~\cite{loshchilov2017decoupled} for
50-200 iterations using a learning rate of 5e-4, a weight decay of 5e-4,
and a cosine scheduler.
During refinement, we inject a small amount of Gaussian noise (\ie, std of 3.5e-2) and apply dropout (\ie, 2.5e-2) to the image
embeddings to stabilize training. 
For few-shot experiments, we follow standard practices and report the average Top-1 accuracy over 3 runs. 
For additional details, refer the supplementary material.

\subsection{Comparisons with State-of-the-Art}

\vspace{0.05in}
\noindent\textbf{Standard Few-shot Image Classification:} As show in \cref{tab:backbones}, LFA outperforms CoOp~\cite{zhou2022learning}
by 1\% on average over the 11 datasets and 
with various visual backbones, with the biggest gains observed on datasets with larger domain gaps, \ie, $\approx 7\%$ on EuroSAT.

\vspace{0.05in}
\noindent\textbf{Base-to-New Generalisation:}
Using 2 mappings, and as shown in \cref{tab:results_generalization_sota}, 
LFA improves the prior best result of ProDA by $2.18\%$ in terms of harmonic
mean, with similar improvements for both base and new classes. Again, the largest gains are observed for datasets with larger
domain gaps, such as UCF101 and EuroSAT.

\begin{table}[ht]
    \tabstyle{1pt}
    \centering
    \caption{\textbf{Base-to-New generalization:} Top-1 acc. for 16-shot per base class.}
    \label{tab:results_generalization_sota}
    \resizebox{\linewidth}{!}{%
    \begin{tabular}{l@{\hspace{0.2cm}}c@{\hspace{0.5cm}}c@{\hspace{0.2cm}}c@{\hspace{0.2cm}}c@{\hspace{0.2cm}}c@{\hspace{0.2cm}}g@{\hspace{0.2cm}}g}
    \toprule
    Dataset & Set & CLIP & CoOp & CoCoOp & ProDA & LFA & $\Delta$ \\
    \midrule
    \multirow{3}{*}{ImageNet}&Base & 72.43 & 76.47 & 75.98 & 75.40 & \textbf{76.89} &        \\
    &New & 68.14 & 67.88 & \textbf{70.43} & 70.23  & 69.36 &        \\
    & H & 70.22 & 71.92 & \textbf{73.10} & 72.72 & 72.93 & \textcolor{Black}{\textbf{-0.17}} \\
    \midrule
    \multirow{3}{*}{Caltech101}&Base & 96.84 & 98.0 & 97.96 & 98.27 & \textbf{98.41} &        \\
    &New & \textbf{94.00} & 89.91 & 93.81 & 93.23  & 93.93 &        \\
    & H & 95.40 & 93.73 & 95.84 & 95.86  & \textbf{96.13} &\textcolor{IndigoPurple}{\textbf{+0.27}}  \\
    \midrule
    \multirow{3}{*}{Pets}&Base & 91.17 & \textbf{93.67} & 95.20 & 95.43  & 95.13 &        \\
    &New & 97.26 & 95.29 & \textbf{97.69} & 97.83  & 96.23 &        \\
    & H & 94.12 & 94.47 & 96.43 & \textbf{96.62}  & 95.68 & \textcolor{Black}{\textbf{-0.94}} \\
    \midrule
    \multirow{3}{*}{Cars}&Base & 63.37 & \textbf{78.12} & 70.49 & 74.70  & 76.32 &        \\
    &New & \textbf{74.89} & 60.40 & 73.59 & 71.20  &  \textbf{74.88} &      \\
    & H & 68.85 & 68.13& 72.01 & 72.91  &  \textbf{75.59} & \textcolor{IndigoPurple}{\textbf{+2.68}} \\
    \midrule
    \multirow{3}{*}{Flowers102}&Base & 72.08 & 97.60 & 94.87 & \textbf{97.70}  & 97.34  &        \\
    &New & \textbf{77.80} & 59.67 & 71.75 & 68.68  & 75.44 &  \\
    & H & 74.83 & 74.06 & 81.71 & 80.66  & \textbf{85.00} & \textcolor{IndigoPurple}{\textbf{+3.29}}  \\
    \midrule
    \multirow{3}{*}{Food101}&Base & 90.10 & 88.33 & 90.70 & 90.30  & \textbf{90.52} &        \\
    &New & 91.22 & 82.26 & 91.29 & 88.57  & \textbf{91.48} &        \\
    & H & 90.66 & 85.19 & \textbf{90.99} & 89.43  & \textbf{91.00} & \textcolor{Black}{\textbf{+0.0}}  \\
    \midrule
    \multirow{3}{*}{Aircraft}&Base & 27.19 & 40.44 & 33.41 & 36.90  & \textbf{41.48} &        \\
    &New & \textbf{36.29} & 22.3 & 23.71 & 34.13 & 32.29 &        \\
    & H & 31.09 & 28.75 & 27.74 & 35.46 & \textbf{36.31} & \textcolor{IndigoPurple}{\textbf{+0.85}} \\
    \midrule
    \multirow{3}{*}{SUN397}&Base & 69.36 & 80.6 & 79.74 & 78.67  & \textbf{82.13} &        \\
    &New & 75.35 & 65.89 & 76.86 & 76.93  & \textbf{77.20} &        \\
    & H & 72.23 & 72.51 & 78.27 & 77.79  & \textbf{79.59} & \textcolor{IndigoPurple}{\textbf{+1.78}} \\
    \midrule
    \multirow{3}{*}{DTD}&Base & 53.24 & 79.44 & 77.01 & 80.67  & \textbf{81.29} &        \\
    &New & 59.9 & 41.18 & 56.0 & 56.48 & \textbf{60.63} &       \\
    & H & 56.37 & 54.24 & 64.85 & 66.44 & \textbf{69.46} & \textcolor{IndigoPurple}{\textbf{+3.02}} \\
    \midrule
    \multirow{3}{*}{EuroSAT}&Base & 56.48 & 92.19 & 87.49 & 83.90 & \textbf{93.40} &        \\
    &New & 64.05 & 54.74 & 60.04 & 66.0 & \textbf{71.24} &        \\
    & H & 60.03 & 68.9 & 71.21 & 73.88  & \textbf{80.83} & \textcolor{IndigoPurple}{\textbf{+6.95}} \\
    \midrule
    \multirow{3}{*}{UCF101}&Base & 70.53 & 84.69 & 82.33 & 85.23  & \textbf{86.97} &        \\
    &New & 77.50 & 56.05 & 73.45 & 71.97  & 77.48 &        \\
    & H & 73.85 & 67.46 & 77.64 & 78.04 & \textbf{81.95} & \textcolor{IndigoPurple}{\textbf{+3.90}} \\
    \midrule
    \multirow{3}{*}{\textbf{Average}}&Base & 69.34 & 82.69 & 80.47 & 81.56 & \textbf{83.62} &        \\
    &New & 74.22 & 63.22 & 71.69 & 72.30  & \textbf{74.56} &        \\
    & H & 71.70 & 71.66 & 75.83 & 76.65  & \textbf{78.83} & \textcolor{IndigoPurple}{\textbf{+2.18}} \\
    \bottomrule
    \end{tabular}
    }
\vspace{-0.35cm}
\end{table}

\vspace{0.05in}
\noindent\textbf{Domain Generalisation:} 
As in base-to-new generalisation, we report the results obtained with two mappings as detailed in \cref{subsec:two-mappings}.
As shown in \cref{tab:img-ood}, LFA outperforms CoOp \cite{zhou2022learning}
on the source domain, while also outperforming CoCoOp \cite{zhou2022conditional} on the target domains,
These results further demonstrate the flexibility of LFA, which can be used to adapt to different domains and settings,
and even under a test-time distribution shift, either on the language or the image side.

\vspace{0.05in}
\noindent\textbf{Standard Action Recognition:} 
\cref{tab:action-recognition-full} shows the obtained action recognition results
when using the full-training set for video-text alignment.
LFA slightly outperforms the video soft-prompting method of \cite{ju2022prompting}
on average, and by a notable margin on HMDB51. However, and different from \cite{ju2022prompting}
that trains two transformer layers on top of the frozen per-frame CLIP features to model the temporal information
in addition to the soft-prompt, LFA matches performances of \cite{ju2022prompting} without any temporal modeling.

\vspace{0.05in}
\noindent\textbf{Few-shot Action Recognition:} similar to the standard setup, and as shown in \cref{tab:action-recognition-fewshot},
LFA largely matches or outperforms the performances of \cite{ju2022prompting} \textit{with no temporal modeling}.

\begin{table*}[ht]
\caption{\textbf{Aligning Disjoint Modalities for Few-shot Classification:} Top-1 acc. for 16-shot per class 
after aligning
the visual and language features of separate (i.e. uni-modal) vision and language encoders. We use 3 self-supervised RN50 vision
encoders 
and the OpenAI embeddings API to access the cpt-text encoder \cite{neelakantan2022text} and generate the class prototypes.
kNN classifier and linear probe results are obtained by training on the visual features.}
\label{tab:selfsup}
\vspace{-0.05in}
\centering
\resizebox{\linewidth}{!}{%
\begin{tabular}{lll*{12}cc}
\toprule
Visual Enc. & Text Enc. & Method & Pets & Flowers102 & Aircraft & DTD & EuroSAT & Cars & Food101 & SUN397 & Caltech101 & UCF101 & ImageNet & Avg. & $\Delta$ \\ 
\midrule
\multirow{3}*{BYOL \cite{grill2020bootstrap}} & \multirow{3}*{OpenAI Emb.} & kNN & 69.24 & 75.84 & 11.63 & 54.77 & 80.58 & 10.88 & 30.13 & 42.04 & 85.89 & 51.12 & 44.57 & 50.61 \\
&& Linear Probe & 76.08 & 82.70 & 13.80 & 62.07 & 87.06 & 14.87 & 37.01 & 47.50 & 90.04 & 59.59 & 48.91 & 56.33 \\
\rowcolor{tabhighlight} 
&& \textbf{LFA}  & {\bf79.84} & {\bf92.81} & {\bf32.03} & {\bf63.83} & {\bf88.82} & {\bf36.98} & {\bf45.78} & {\bf53.72} & {\bf92.2} & {\bf66.82} & {\bf54.04} & {\bf64.26} & {\bf\textcolor{IndigoPurple}{+7.93}} \\
\midrule
\multirow{3}*{BarlowTwins~\cite{zbontar2021barlow}} & \multirow{3}*{OpenAI Emb.} & kNN & 68.73 & 79.87 & 14.79 & 54.73 & 81.66 & 12.01 & 30.71 & 41.35 & 84.07 & 49.07 & 41.44 & 50.77 \\
&& Linear Probe & 75.78 & 85.14 & 16.92 & 62.53 & 87.92 & 15.89 & 37.4 & 45.94 & 88.93 & 56.6 & 45.01 & 56.19 \\
\rowcolor{tabhighlight} 
&& \textbf{LFA}  & {\bf79.39} & {\bf93.24} & {\bf35.02} & {\bf63.71} & {\bf89.65} & {\bf41.29} & {\bf46.02} & {\bf52.66} & {\bf91.70} & {\bf64.95} & {\bf51.34} & {\bf64.45} &  {\bf\textcolor{IndigoPurple}{+8.26}} \\
\midrule
\multirow{3}*{MoCo v3~\cite{chen2021empirical}} & \multirow{3}*{OpenAI Emb.} & kNN &  78.13 & 80.63 & 18.66 & 54.39 & 81.74 & 14.56 & 32.10 & 44.23 & 91.13 & 55.36 & 50.10 & 54.64 \\
&& Linear Probe & 83.81 & 86.97 & 20.59 & 61.94 & 88.61 & 18.83 & 39.47 & 49.24 & 93.52 & 63.34 & 53.55 & 59.99 \\
\rowcolor{tabhighlight} 
&& \textbf{LFA}  & {\bf85.18} & {\bf93.36} & {\bf39.75} & {\bf62.71} & {\bf89.13} & {\bf45.36} & {\bf46.15} & {\bf54.22} & {\bf94.25} & {\bf68.21} & {\bf57.57} & {\bf66.90} & {\bf\textcolor{IndigoPurple}{+6.91}} \\
\bottomrule
\end{tabular}
}
\vspace{-0.1cm}
\end{table*}

\subsection{Ablation studies}
\label{sec:ablations}

In this section, we (1) ablate the impact of the proposed closed-form relaxed initialisation,
(2) compare the new re-ranking loss with a series of baselines, and (3) analyse the effect of
the number of image crops and of template-based text prompting for class prototyping. Moreover,
to showcase the generalisability of our approach, (4) we explore our method's behaviour on disjoint models,
where the vision and text encoder are trained from separate sources, (5) the effectiveness of LFA with V-L models other than
CLIP, (6) the results sensitivity to the choice of $\beta$. Finally, (7) we report results for the
unsupervised variant of our method. See supplementary material for some additional ablations.

\vspace{0.05in}
\noindent\textbf{Effect of closed-form initialisation and refinement:} The proposed $\beta$-Procrustes regularization
significantly reduces overfitting (\cref{tab:ablation-beta-procrustes}) and provides a notably better starting point
for the refinement process (\cref{tab:ablation-procrustes-refine}) across all datasets.

\vspace{0.05in}
\noindent\textbf{ARerank vs other losses:} In \cref{tab:ablation-refinement-loss} we compare the proposed ARerank loss with a series of baselines,
out of with the CSLS loss is the closest conceptually to ours. As the results show, we outperform CSLS across all datasets by up to $4\%$.
Moreover, we improve between 1.6\% (for Imagenet) and $5\%$ (for FGVC) on top of the standard contrastive loss. 

\vspace{0.05in}
\noindent\textbf{Effect of the number of image crops and prototype templating:} As show in
\cref{tab:ablation-augmentation}, LFA is largely invariant to the exact number
of crops per image used for training, showing strong results with a single crop (\cref{tab:ablation-augmentation}).
Additionally, and similar to~\cite{bulat2022language,zhou2022conditional}, the results show that the usage of prompting
templates for constructing the class prototypes to be beneficial.

\begin{table}[t]
    \caption{\textbf{LFA and Soft-Prompts:} Top-1 acc. on ImageNet for 16-shot per class when
    using trained soft-prompts (\eg, CoOp \cite{zhou2022learning}) to generate the  class prototypes.}
    \vspace{-0.15in}
    \label{tab:ablation-coop}
    \vspace{5pt}
    \centering
    \resizebox{\linewidth}{!}{%
    \begin{tabular}{l*{4}c}
    \toprule
    Backbone & RN50 & RN101 & ViT-B/32 & ViT-B/16 \\ 
    \midrule
    CoOp \hspace{1.0in} & 62.62 & 66.45 & 66.41 & 71.62 \\
    LFA & 63.65 & 67.16 & 67.63 & 72.61 \\
    \rowcolor{tabhighlight} LFA + CoOp & \textbf{63.78} & \textbf{67.72} & \textbf{67.85} & \textbf{73.10} \\
    \bottomrule
    \vspace{-0.2in}
    \end{tabular}
    }
\end{table}

\vspace{0.05in}
\noindent\textbf{LFA and soft-prompts:} To further show the complementarity and flexibility of LFA, we use a set of pre-trained
soft-prompts (\ie, CoOp \cite{zhou2022learning}) to obtain the class prototypes. Then we proceed with the LFA procedure.
As shown in \cref{tab:ablation-coop}, LFA can also be coupled with soft-prompts for additional improvements. 

\vspace{0.05in}
\noindent\textbf{Other V-L models:}
Given the black-box nature of LFA, it can be used as 
is with other V-L models and with similar gains in performance as CLIP.
\cref{tab:ablation-other-models} shows the results obtained with LFA
when using other V-L models further confirming the generality and flexibility of LFA.

\begin{table}[t]
  \caption{\textbf{LFA with various V-L models:} Top-1 acc. on ImageNet for 16-shot per class when
  using features from V-L models other than CLIP.}
  \vspace{-0.1in}
  \label{tab:ablation-other-models}
  \centering
  \resizebox{\linewidth}{!}{%
  \begin{tabular}{l*{3}c}
  \midrule
  Method\hspace{1.4in} & ALIGN~\cite{jia2021scaling} & FLAVA~\cite{jia2021scaling} & AltCLIP~\cite{jia2021scaling} \\ 
  \midrule
  Zero-shot & 64.4 & 54.6 & 73.5 \\
  \rowcolor{tabhighlight} LFA (ImageNet 16-shot) & \textbf{69.8} (+5.4) & \textbf{61.1} (+6.5) & \textbf{79.1} (+5.6) \\
  \midrule
  \vspace{-0.2in}
  \end{tabular}
}
\end{table}

\vspace{0.05in}
\noindent\textbf{Sensitivity to the choice of $\beta$:} 
While it is beneficial to tune $\beta$ for each dataset using cross-validation,
\cref{tab:ablation-beta} shows that the final results remain robust to the choice of $\beta$, and
setting $\beta \in [0.6, 0.9]$ results in similar performances.

\begin{table}[t]
  \caption{\textbf{Sensitivity to $\beta$:} Top-1 acc. on ImageNet for 16-shot per class with different $\beta$ values.}
  \vspace{-0.1in}
  \label{tab:ablation-beta}
  \centering
  \resizebox{\linewidth}{!}{%
  \begin{tabular}{l*{10}c}
  \midrule
  $\beta$ & 0.0 & 0.1 & 0.2 & 0.3 & 0.4 & 0.5 & 0.6 & 0.7 & 0.8 & 0.9 \\ 
  \midrule
  LFA & 71.04 & 71.39 & 71.67 & 71.87 & 72.08 & 72.26 & 72.45 & \textbf{72.54} & \textbf{72.56} & \textbf{72.54} \\
  \midrule
  \vspace{-0.2in}
  \end{tabular}
}
\end{table}

\vspace{0.05in}
\noindent\textbf{Performance analysis for disjoint modalities:} Our approach is modality, domain, and architecture agnostic.
Moreover, it doesn't require access to the weights, only to the produced features. To showcase this,
we introduce a new evaluation setting in which the visual and text features are produced by disjoint models, that never
interacted during training. Either one or both modalities can be sourced from behind-the-wall (\ie, blackbox) models.
For this experiment, we consider 3 RN50 pre-trained visual backbones: BYOL, BarlowTwins and MoCo v3. As we do not require
access to the model, we use the OpenAI embeddings API\footnote{\url{https://platform.openai.com/docs/api-reference/embeddings/}}
to get the text embeddings from the cpt-text encoder \cite{neelakantan2022text} and generate the class prototypes. 
After an initial random projection to project the image features into the 1536-d space and match the dimensionality of
text features, we proceed with the supervised LFA procedure as with CLIP experiments.
In a few-shot setting, alongside our method, we consider 2 baselines that also operate the frozen
visual features: kNN and linear eval (see supplementary material for the hyper-parameters used).
As the results from \cref{tab:selfsup} show, our method (1) reaches performance comparable with that of
aligned models (\ie, CLIP) and (2) outperforms both kNN and linear eval by a large margin.

\vspace{0.05in}
\noindent \textbf{Unsupervised LFA:} As presented in \cref{ssec:overall_lfa}, LFA can be adapted for label-free training.
\cref{tab:unsup-alignment} shows that U-LFA improves by up to 7\% on top of zero-shot CLIP.

\section{Conclusions}

In this work we proposed LFA, the first black box few-shot adaptation method for V-L models,
that uses pre-computed image and text features without accessing the model's weights. Other advantages of
LFA include fast training, applicability to both supervised and unsupervised setting, and even application for
aligning image and text features computed from uni-modal models. Thanks to the use of precomputed features, we
hope that LFA will enable few-shot adaptation of very large V-L foundation models that would otherwise be
impossible to adapt or even deploy. 

\appendix

\section{Appendix}

\begin{table*}[t]
\caption{\textbf{Few-shot Classification:} the obtained average Top-1 test acc. on 11 classification datasets
with CoOp \cite{zhou2022learning}, Linear Probe, and the proposed LFA, with either 4, 8 or 16-shot per class and with RN50 as the visual encoder.}
\label{tab:n-few-shot}
\vspace{-0.1in}
\centering
\resizebox{\linewidth}{!}{%
\begin{tabular}{cl*{12}cc}
\toprule
N-shot & Method & Pets & Flowers102 & Aircraft & DTD & EuroSAT & Cars & Food101 & SUN397 & Caltech101 & UCF101 & ImageNet & Avg. & $\Delta$ \\ 
\midrule
& CLIP RN50 & 85.77 & 66.14 & 17.28 & 42.32 & 37.56 & 55.61 & 77.31 & 58.52 & 86.29 & 61.46 & 58.18 & 58.77 \\
\midrule
\multirow{3}*{4} &
Linear Probe & 56.35 & 84.80 & 23.57 & 50.06	& 68.27	& 48.42	& 55.15	& 54.59	& 84.34 & 62.23	& 41.29 & 57.19\\
& CoOp & {\bf86.06}	& 86.52	& 22.02	& 52.72	&  {\bf70.93} &  {\bf61.62}	& {\bf72.64} & 63.67	& 88.53	& 67.06 &  {\bf59.96} & 66.52\\
\rowcolor{tabhighlight} &
\textbf{LFA} & 82.21 &  {\bf88.28} &  {\bf24.15} &  {\bf54.51} & 68.76 & 60.69 & 71.81 &  {\bf65.50} & {\bf88.88} & {\bf69.25} & 58.36 &  {\bf66.58} & {\bf+0.06} \\
\midrule
\multirow{3}*{8} &
Linear Probe & 65.94 &	92.00 &	29.55 &	56.56 &	{\bf76.93} &	60.82 &	63.82 &	62.17 &	87.78 &	69.64 &	49.55 &	64.98 \\
& CoOp & 83.58 & 91.81 & 28.18 &	59.14 &	77.65 &	67.32 &	72.04 &	65.64 &	90.33 &	72.74 &	{\bf62.04} &	70.04 \\
\rowcolor{tabhighlight} &
\textbf{LFA} & {\bf84.93} & {\bf92.62} & {\bf30.17} & {\bf60.54} & 77.36 & {\bf67.60} & {\bf74.47} & {\bf68.54} & {\bf91.33} & {\bf73.33} & 61.36 & {\bf71.10} & {\bf\textcolor{IndigoPurple}{+1.06}} \\
\midrule
\multirow{3}*{16} &
Linear Probe & 76.42 &	{\bf94.95} & {\bf36.39} &	63.97 &	82.76 &	70.08 &	70.17 &	67.15 &	90.63 &	73.72 &	55.87 &	71.10 \\
& CoOp & 86.16 & 94.80 & 32.29 &	63.16 &	83.55 &	73.27 &	74.46 &	69.12 &	91.62 &	75.29 &	63.08 &	73.35 \\
\rowcolor{tabhighlight} &
\textbf{LFA} & {\bf86.75} & 94.56 & 35.86 & {\bf66.35} & {\bf84.13} & {\bf73.58} & {\bf76.32} & {\bf71.32} & {\bf92.68} & {\bf77.00} & {\bf63.65} & {\bf74.75} & {\bf\textcolor{IndigoPurple}{+1.40}} \\
\bottomrule
\end{tabular}
}
\end{table*}

\begin{table*}[t]
    \caption{\textbf{Aligning Disjoint Modalities:} we show the obtained average Top-1 acc. on 11 image classification
    datasets for 8- and 16-shot per class and with either a single center crop or five crops as data augmentation.}
    \label{tab:self-sup-supmat}
    \vspace{-0.05in}
    \centering
    \resizebox{0.95\linewidth}{!}{%
    \begin{tabular}{llcccggcccggcccgg}
    \toprule
    &&& \multicolumn{4}{c}{BYOL} && \multicolumn{4}{c}{BarlowTwins} && \multicolumn{4}{c}{MoCo v3} \\
    \cmidrule{4-7}  \cmidrule{9-12}  \cmidrule{14-17} 
    N-shot & Crops && kNN & Lin. Probe & LFA & $\Delta$ &&  kNN & Lin. Probe & LFA & $\Delta$ &&  kNN & Lin. Probe & LFA & $\Delta$ \\
    \midrule
    16 & 1 && 50.61 & 56.33 & {\bf64.26}  & {\bf\textcolor{IndigoPurple}{+7.93}} && 50.77 & 56.19 & {\bf64.45} & {\bf\textcolor{IndigoPurple}{+8.26}} && 54.64 & 59.99 & {\bf66.90} & {\bf\textcolor{IndigoPurple}{+6.91}} \\
    16 & 5 && 51.69 & 61.24 & {\bf64.48}  & {\bf\textcolor{IndigoPurple}{+3.24}} && 51.64 & 61.49 & {\bf64.91} & {\bf\textcolor{IndigoPurple}{+3.41}} && 55.7 & 64.79 & {\bf67.23} & {\bf\textcolor{IndigoPurple}{+2.44}} \\
    8  & 1 && 44.27 & 51.09 & {\bf58.15}  & {\bf\textcolor{IndigoPurple}{+7.05}} && 43.86 & 50.70 & {\bf58.08} & {\bf\textcolor{IndigoPurple}{+7.37}} && 48.04 & 54.77 & {\bf60.93} & {\bf\textcolor{IndigoPurple}{+6.15}} \\
    8  & 5 && 47.07 & 55.69 & {\bf58.31}  & {\bf\textcolor{IndigoPurple}{+2.62}} && 47.02 & 55.71 & {\bf58.52} & {\bf\textcolor{IndigoPurple}{+2.81}} && 51.11 & 59.16 & {\bf61.00} & {\bf\textcolor{IndigoPurple}{+1.84}} \\
    \bottomrule
    \end{tabular}
    }
    \end{table*}

\subsection{Additional Ablations}

\noindent\textbf{Few-shot Image Classification} 
\cref{tab:n-few-shot} shows the obtained results with different numbers of support examples per class,
\ie 4, 8 and 16-shot training data per-class. Similar to the results presented in \cref{sec:experiments},
LFA either matches (\ie for 4-shot) or outperforms (\ie for 8 and 4-shot) soft-prompting.

\vspace{0.1in}
\noindent\textbf{Aligning Uni-modal Models:}
In \cref{tab:selfsup}, we reported the results when aligning
self-supervised vision models (\ie, BYOL, BarlowTwins, and MoCo v3) and the cpt-text encoder accessed via OpenAI's embeddings API for
16-shot per class and with a single center crop. In \cref{tab:self-sup-supmat}, 
we show the obtained average accuracy on the 11 image datasets for 8- and 16-shot per class,
and with either a single center crop or five crops.
Overall, LFA outperforms kNN and linear probe classifiers, and by a wide margin when 
the labeled data is scarce (\ie, a single crop).
Note that another benefit of LFA is the possibility of using data augmentation in
the language domain, \eg, prompt ensembling, instead of a single standard prompt,
\ie, ``\texttt{a photo of a \{{cls name}\}}'', which can further boost the performances.

\subsection{Experimental Details}
Overall, the training procedure of LFA remains as detailed in \cref{alg:LFA}.
After the $\beta$-Procrustes initialisation
of the mapping $\mathbf{W}$, we refine it using ARerank loss for
given number of iteration (\ie to be detailed on a per-dataset basis)
using AdamW with a learning rate of 5e-4, a weight decay of 5e-4 and a cosine scheduler that decreases the learning
rate to 1e-7 by the end of training. For most experiments, and unless noted otherwise,
we add a small amount of Gaussian noise (\ie std of 3.5e-2)
and apply dropout (\ie probability of 2.5e-2) to the image embeddings to avoid overfitting
and make the training robust to the choice of
the number of refinement steps. Next, we will present the different
experimental details on a per-setting basis.
For each experiment, the class prototypes are generated by inserting the class name in the standard
templates~\cite{zhou2022learning} \eg, ``\texttt{a photo of a \{{cls name}\}}''
for image tasks and ``\texttt{a video frame of a person \{{action type}\}}'', in order to give
a better initialisation of the prototypes and facilitate the image-text alignment.
For all few-shot results, we report the average accuracy over 3 runs.

\subsubsection{Standard Few-shot Image Classification} 
In this setting, we set $\beta$ based on a 3-fold cross validation on the training set
with a 20-70 validation-train split. We select the $\beta$ with the highest average validation accuracy
across the 3-folds and from a set of values in the range [0.0, 1.0] with a step of 0.05.
We generate 5 crops for each training image of size 224x224, \ie four corner crops and a central crop,
and use their features for training. While the training is robust to the number of refinement
steps, we noticed that the better the orthogonal and $\beta$-Procrustes initialisation results,
and less number of steps needed to obtain the optimal mapping. 
\cref{tab:refine-steps-standard} shows the number of refinement for each image classification
dataset.

\subsubsection{Base-to-New (Zero-Shot) Recognition} 
For base-to-new experiments, we found that $\beta=0.9$
performs well on most dataset without requiring a cross-validation step.
Similar to the standard few-shot setting, we train with 5 crops
and refine the mapping for up to 100 iterations. See 
\cref{tab:refine-base-new} for the number of refinement steps for each dataset.

\subsubsection{Domain Generalisation:} 
For domain generalisation experiments, we set $\beta=0.9$
and train with 5 crops for 200 refinement iterations.

\subsubsection{Action Recognition:} 
For few-shot action recognition experiments, we set $\beta=0.9$
and train with a single center crop. For UCF101, no dropout is used
and the Gaussian noise is reduced to 2.5e-2, and we conduct 300 refinement steps.
For HMDB51, conduct 100 refinement steps.
When using the whole training set for alignment, we use the same setup as the few-shot
setting, but we conduct 500 refinement steps for UCF101 and 300 for HMDB51.

\subsubsection{Aligning Disjoint Modalities} 
For the alignment of the features of uni-modal models, we use 3 self-supervised RN50
visual encoders: BYOL, BarlowTwins and MoCo v3, in addition to the cpt-text 
encoder as our language encoder.
After extracting the features, we first conduct a Gaussian random projection implemented using \texttt{scikit-learn}
\cite{pedregosa2011scikit}
to reduce the dimensionality of the visual features from 2048 to 1536 to
match those of the text encoder. Then we proceed as in the standard few-shot classification
setup, by first finding $\beta$ with cross-validation, initializing the mapping with 
$\beta$-Procrustes, then refining it for 700-800 iterations using 5 crops.

In terms of the kNN and linear probe baselines, we follow the practical recommendations
of \cite{tian2020rethinking}, and train the classifiers on the $\ell_2$ normalized and forzen 
visual features (\ie the original 2048-d features). For the kNN classification, we use 16
neighbors for training, as for the linear probe, we follow \cite{tian2020rethinking} and 
use the multinomial logistic regression implementations of \texttt{scikit-learn}, and train with an 
$\ell_2$ penalty and the LBFGS solver. Similar to LFA, all baselines were trained with 5 crops.

\begin{table}[H]
  \caption{\textbf{Refinement Steps for Standard Few-shot Classification:} we specify 
  the number of refinement steps on a per-dataset basis.}
  \vspace{-0.05in}
  \label{tab:refine-steps-standard}
  \vspace{5pt}
  \centering
  \resizebox{0.95\linewidth}{!}{%
  \begin{tabular}{lc}
  \toprule
  Datasets & Nbr. refinement steps \\ 
  \midrule
  Cars & 2000  \\
  Caltech101, DTD, Aircraft & 1000 \\
  EuroSAT, Food101, ImageNet, UCF101 & 200 \\
  Flowers, SUN397 & 100 \\
  Pets & 30 \\
  \bottomrule
  \end{tabular}
  }
\end{table}
\begin{table}[H]
  \caption{\textbf{Refinement Steps for Base-to-New (Zero-Shot) Recognition:} we specify 
  the number of refinement steps on a per-dataset basis.}
  \vspace{-0.05in}
  \label{tab:refine-base-new}
  \vspace{5pt}
  \centering
  \resizebox{0.95\linewidth}{!}{%
  \begin{tabular}{lc}
  \toprule
  Datasets & Nbr. refinement steps \\ 
  \midrule
  Caltech101, DTD, UCF101, ImageNet & 100 \\
  EuroSAT, Food101, Flowers, Cars, SUN397 & 50 \\
  Caltech101 & 40 \\
  Pets & 10 \\
  \bottomrule
  \end{tabular}
  }
\end{table}

\clearpage
{\small
\bibliographystyle{ieee_fullname}
\bibliography{references}
}

\end{document}